\pdfoutput=1
\documentclass[11pt]{article}

\usepackage[preprint]{acl}

\usepackage{times}
\usepackage{latexsym}

\usepackage[T1]{fontenc}
\usepackage[utf8]{inputenc}

\usepackage{microtype}

\usepackage{inconsolata}

\usepackage{graphicx}

\usepackage{booktabs}
\usepackage{linguex}
\usepackage{subcaption}
\usepackage{multirow}
\usepackage[inline]{enumitem}
\usepackage{parskip}
\usepackage[skins,breakable]{tcolorbox}
\usepackage{float}

\definecolor{burntorange}{HTML}{bf5700}
\newtcolorbox[list inside=prompt,auto counter,number within=section]{prompt}[1][]{
    colbacktitle=burntorange,
    colframe=burntorange,
    fontupper=\footnotesize,
    boxsep=5pt,
    left=0pt,
    right=0pt,
    top=0pt,
    bottom=0pt,
    boxrule=1pt,
    enhanced, 
    breakable,
    skin first=enhanced,
    skin middle=enhanced,
    skin last=enhanced,
    #1,
}

\title{\emph{Dark \& Stormy}:\\ Modeling Humor in Sentences from the Bulwer-Lytton Fiction Contest}

\author{Venkata S Govindarajan \\
  Department of Computer Science, \\
  Ithaca College \\
  \texttt{vgovindarajan@ithaca.edu} \\\And
  Laura Biester \\
  Department of Computer Science, \\
  Middlebury College \\
  \texttt{lbiester@middlebury.edu} \\}

\begin{document}

\setlength{\Exlabelwidth}{0.8em}
\setlength{\Exlabelsep}{0.8em}
\setlength{\SubExleftmargin}{1em}
\setlength{\Extopsep}{0.5\baselineskip}

\maketitle

\begin{abstract}
Textual humor is enormously diverse and computational studies need to account for this range, including \textbf{intentionally bad humor}. In this paper, we curate and analyze a novel corpus of sentences from the Bulwer-Lytton Fiction Contest to better understand ``bad'' humor in English. Standard humor detection models perform poorly on our corpus, and an analysis of literary devices finds that these sentences combine features common in existing humor datasets (e.g., puns, irony) with metaphor, metafiction and simile. Large language models (LLMs) prompted to synthesize contest-style sentences imitate the form but exaggerate the effect by over-using certain literary devices and including far more novel adjective-noun bigrams than human writers.
\end{abstract}

\section{Introduction}
\label{sec:intro}
Humor is a uniquely human trait, and its interpretation involves language and social understanding, creativity,  and common-sense reasoning~\citep{Brock2017ModellingTC, martin2018psychology, attardo2020linguistics}, making it an attractive challenge for Computational Linguistics. Prior work has explored a range of humor-related tasks, including humor detection~\citep{taylor2004computationally,mihal2006}, generation~\citep{hessel-etal-2023-androids}, and even editing~\citep{horvitz-etal-2024-getting}. However, most available humor datasets focus on familiar forms such as satirical headlines, puns, or short one-liners, leaving other varieties underexplored. In this paper, we introduce a novel dataset for an understudied form of textual humor --- intentionally bad humor --- through sentences drawn from the \textbf{Bulwer-Lytton Fiction Contest\footnote{\url{https://www.bulwer-lytton.com/}}} (BLFC).

The BLFC was an annual competition (1982--2024) founded by Professor Scott Rice at San Jose State University, later co-run with his daughter Elizabeth J. Rice. Inspired by the notorious opening line of Edward Bulwer-Lytton's 1830 English novel \emph{Paul Clifford}, it invited participants to craft ``an atrocious opening sentence to the worst novel ever written.'' Each year, the organizers selected 50--60 entries to highlight from thousands of submissions. Figure~\ref{fig:example-sent} is an example entry, with dominant literary devices highlighted (see Appendix~\ref{app:bl-exs} for more examples). \textbf{BLFC sentences thus derive their humor from being  crafted to be intentionally bad}.

\begin{figure}
    \centering
    \includegraphics[width=\linewidth]{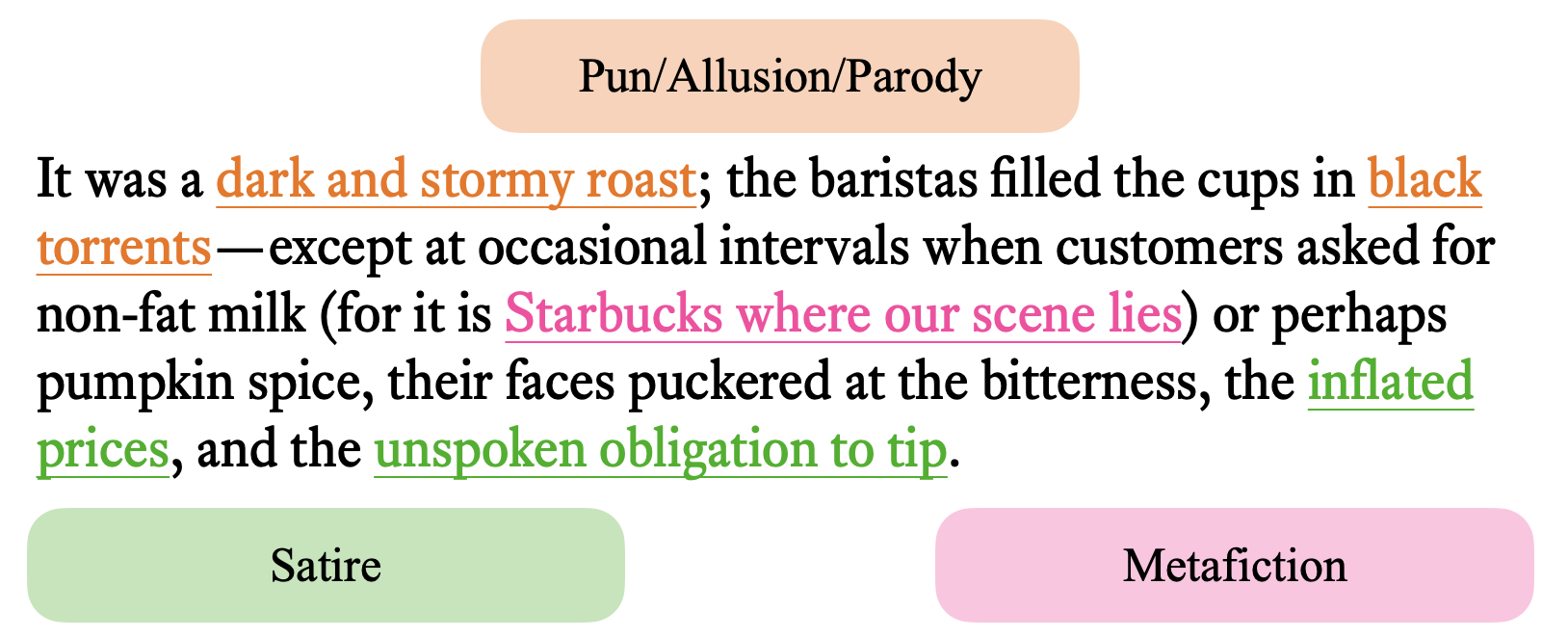}
    \caption{An entry from the Bulwer-Lytton contest, with the key literary devices highlighted.}
    \label{fig:example-sent}
    \vspace{-1\baselineskip}
\end{figure}

In this paper, we build and analyze a unique resource of intentionally bad humor with selected entries from the BLFC. Our goals are twofold: (1) to show that the form of humor in these sentences differs significantly from standard humor datasets, and (2) to examine how Large Language Model (LLM) generated Bulwer-Lytton sentences compare with human-written ones. 

We find that Bulwer-Lytton sentences differ sharply from the texts in standard humor datasets. Humor detection models perform poorly on them, they make greater use of literary devices like metaphor, metafiction, and simile, and they contain many more \emph{semantically deviant} (novel) adjective-noun expressions. Moreover, while LLM-generated sentences can imitate the form of Bulwer-Lytton humor, they consistently exaggerate the effect across all of our analyses. We share our data and code online at \href{https://github.com/venkatasg/bulwer-lytton}{\ttfamily github.com/venkatasg/bulwer-lytton}.

\section{Data}
\label{sec:data}
\paragraph{Bulwer-Lytton sentences} With permission from the contest organizers, we scrapped all entries on the contest website between 1996 and 2024. The organizers list entries for every year by genre, and also list a grand prize winner, runner-up, and other ad-hoc categories --- meta-data that we also collect and associate with each entry. Thus, we compile the Bulwer-Lytton (henceforth abbreviated to \textbf{BL}) dataset comprising 1778 sentences over 29 years. 

\begin{table}[t]
    \centering
    \small
    \begin{tabular}{lrr}
        \toprule
        Dataset & Avg. token len (S.D.) & \# Sentences\\\midrule
         BL & 70.5 (23.1) & 1778\\
         novel-openings & 24.5 (30.8) & 7901\\
         combo-humor & 19.0 (8.9) & 5416\\
         PotD & 14.3 (5.1) & 493 \\
         GPT-5 & 73.7 (6.7) & 1000\\
         DeepSeek & 94.1 (26.6) & 1000\\
         GPT-4.1 & 81.1 (15.5) & 1000\\
         \texttt{gpt-oss-120b} & 95.4 (27.7) & 1000\\
         \bottomrule
    \end{tabular}
    \caption{Average token length (with standard deviation in parentheses) per sentence in each of our datasets. Sentences tokenized with \texttt{gemma-3-1b-pt}.}
    \vspace{-\baselineskip}
    \label{tab:stats}
\end{table}

\paragraph{Synthetic BL sentences} While LLM abilities in humor understanding, generation, and manipulation have been uneven~\citep{hessel-etal-2023-androids, horvitz-etal-2024-getting, cocchieri-etal-2025-call}, they excel at instruction following~\citep{ouyang2022training} and \emph{imitating obscure textual forms}~\citep{reif-etal-2022-recipe}. In this work, we conduct a preliminary comparison of LLM-generated and human-written BL sentences to understand how prompting captures---or distorts---the style of intentionally bad humor.

We use a minimal one-shot prompt (see Appendix~\ref{app:prompt}) describing the contest and its rules taken from the website, along with the original BL sentence. We generate and analyze 1000 BL sentences from \texttt{DeepSeek-V3.1}~\citep{deepseekai2024deepseekv3technicalreport} and OpenAI's \texttt{gpt-5-2025-08-07}~\citep{gpt5} in this paper. We also present results from \texttt{gpt-4.1-2025-04-14}~\citep{gpt41} and \texttt{gpt-oss-120b}~\citep{openai2025gptoss120bgptoss20bmodel} in the appendix.

\paragraph{Baseline datasets}
We aim to capture how BL sentences differ from two other types of texts: first sentences to novels and texts used in other studies of humor. We study first sentences from a crowdsourced list of first sentences to novels (henceforth referred to as \textbf{novel-openings}).\footnote{\url{https://github.com/janelleshane/novel-first-lines-dataset}} The crowd-sourced submissions range from classics (e.g., Jane Austen, Charles Dickens) to modern bestsellers (e.g., Terry Pratchett, Nora Roberts) to sentences submitted by their amateur creators.

As BL sentences exemplify a genre of humor that has been understudied in NLP, we also want to see how they differ from texts that exemplify other genres of humor. We draw from humor datasets compiled by \citet{baranov-etal-2023-told}, and focus on the combined dataset (henceforth referred to as \textbf{combo-humor}) which is a diverse collection of different types of humor. It incorporates puns, satire, ShortJokes from humorous Reddit posts \cite{chen-soo-2018-humor} and jokes from Twitter/ShortJokes drawn from \citet{meaney-etal-2021-semeval}. When considering pun-specific models and BL sentences labeled ``Vile Puns,'' we incorporate Pun of the Day jokes (henceforth referred to as \textbf{PotD}) which were collected by \citet{yang-etal-2015-humor}. We focus on the positive instances from the test split unless otherwise stated.

Table~\ref{tab:stats} shows that BL sentences (human-written and synthetic) \textbf{are substantially longer than typical jokes or novel openings}. We explore how this verbosity contributes to their distinctive form of ``bad humor'' in following sections.

\section{Preliminary Analysis}
\label{sec:analysis}
\subsection{Humor Detection}

\begin{figure*}[t!]
    \centering
    \includegraphics[width=\textwidth]{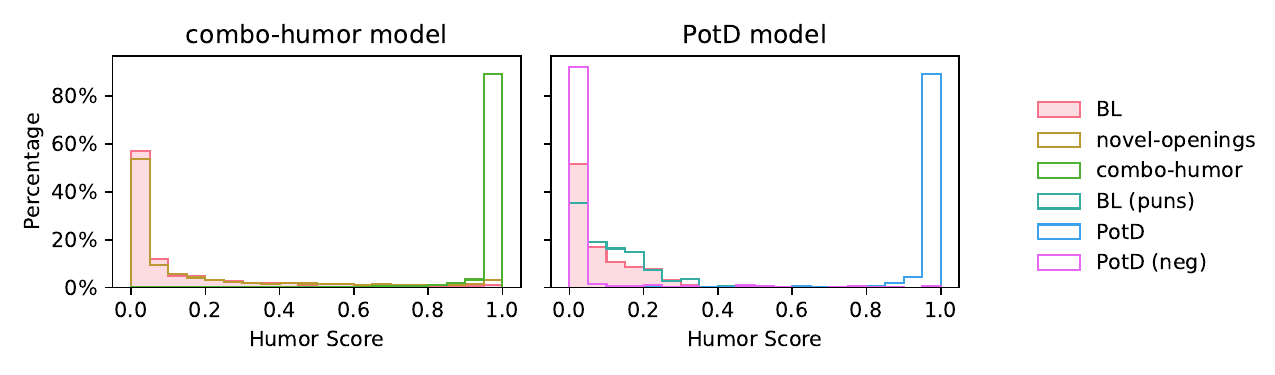}
    \vspace{-2\baselineskip}
    \caption{Comparing humor detected in BL sentences to novel-openings/combo-humor with the combined humor model (L). Comparing humor detected in BL sentences with the pun subset of BL/PotD with the pun model (R).}
    \label{fig:hd-all}
    \vspace{-0.5\baselineskip}
\end{figure*}

In this section, we examine whether existing humor detection models recognize BL sentences as humorous. We use the best performing RoBERTa-based model trained by \citet{baranov-etal-2023-told} on combined humor datasets, as well as their model to detect puns.% We report the mean prediction from the models over 5 random seeds.% and to detect satire (trained on \cite{west2019reverse}).

We analyze humor by plotting a histogram of the continuous scores (0--1) from the models for each dataset, using the mean over 5 random seeds. We find that BL sentences are not identified as humorous; using the combined model, the humor scores are in line with novel-openings (Figure~\ref{fig:hd-all}, left), and results were similar on synthetic BL sentences. Meanwhile, the in-domain sentences from combo-humor (the test split of the dataset the model was trained on) have very high humor scores. 

We also consider whether models can recognize specific types of humor (puns) in out-of-domain text by using a humor detection model trained on PotD and analyzing a subset of the BL sentences that were labeled with the genre ``Vile Puns.'' We find that BL sentences have slightly lower scores than the subset of the sentences that are identified as puns (Figure~\ref{fig:hd-all}, right). However, while both have scores that on average exceed the scores on the negative examples from the PotD test set, they are nowhere near the scores on the positive examples from the in-domain PotD data. Closer inspection reveals that puns are often \emph{one of multiple features} that are employed to make a BL sentence humorous, whereas in the PotD dataset, the examples are typically short sentences which are only funny due to a pun. Our dataset \textbf{demonstrates a failure of pun detection models trained on short jokes to recognize the use of puns in a longer context}.

\subsection{Literary Devices}
\label{subsec:literary}

To further analyze how BL sentences differ from standard humor datasets, we automatically extract sets of literary devices that are used in the sentences using a prompt-based framework adapted from TopicGPT~\cite{pham-etal-2024-topicgpt}. Compared to traditional topic modeling algorithms such as LDA~\cite{blei2003latent}, we find that this method is able to extract higher-level features capturing the \emph{style} of writing rather than content. This method provides structured output, contrary to free-text humor explanations generated by LLMs in prior work \cite{loakman-etal-2025-comparing}.

We add humor-related seed topics and rework the prompts to refer generically to ``features'' rather than ``topics'' to better fit the type of label we hope to extract. We execute the generation and alignment processes with GPT-4.1 using 300 sentences with 100 sampled from each of the BL dataset, the combo-humor dataset (to capture features related to humor), and the novel-openings dataset (to capture features common in first sentences). This process results in a set of eight features, and we validate that the features align with the text using an intruder task (see Appendix~\ref{app:topics}).

\begin{figure}
    \centering
    \includegraphics[width=\linewidth]{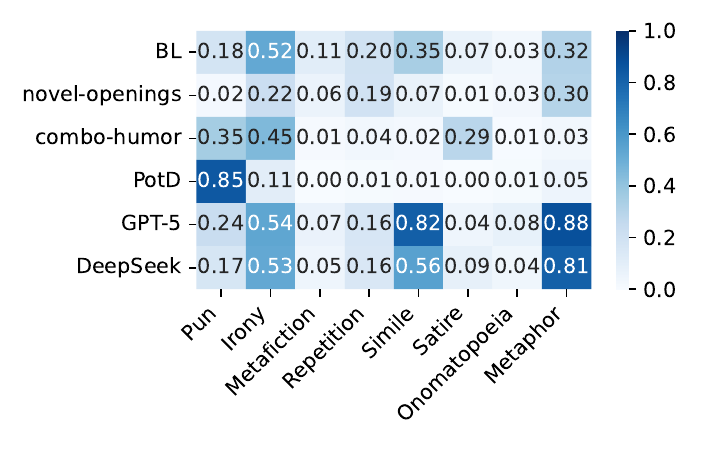}
    \caption{Literary device presence across datasets.}
    \label{fig:topic-v-baseline}
    % \vspace{-\baselineskip}
\end{figure}

After extracting features, we run the assignment framework on all BL sentences (including synthetic) and up to 1000 randomly sampled instances from each baseline dataset. Figure~\ref{fig:topic-v-baseline} demonstrates how the literary devices in the BL dataset differ from the other datasets. We find that BL sentences clearly exceed the baselines in irony, metafiction, and simile. They fall between the crowdsourced novel openings and existing humor datasets in their use of puns and satire. Furthermore, we observe a tendency of the synthetic BL sentences to overuse certain literary devices including simile, metaphor, and (to a lesser extent) onomatopoeia.

\begin{figure*}
    \centering
    \includegraphics[width=\textwidth]{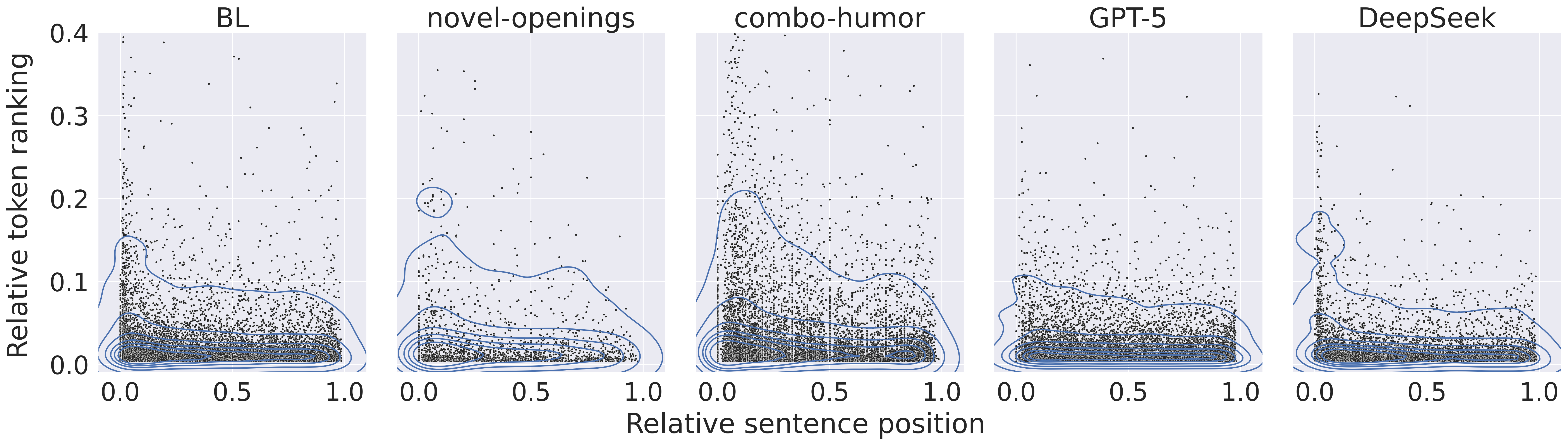}
    \vspace{-\baselineskip}
    \caption{Relative rank of high-surprisal tokens plotted against their relative sentence position for our 5 datasets.}
    \label{fig:surprisal}
    % \vspace{-0.5\baselineskip}
\end{figure*}

% Plots:
% - comb model BL vs good vs Crowd
% - comb model BL vs comb vs unfun vs potd
% - potd model BL vs BL-puns vs potd

% However, the scores are highest for Bulwer-Lytton sentences for two specific types of humor that we would expect to see more commonly in those sentences: puns and satire.\todo{Do we want to frame around comparing to both other forms of humor and other first sentences?}

\section{Incongruity in BL sentences}
\label{sec:exps}
\subsection{Surprisal}
\label{subsec:surprisal}

The incongruity and semantic script theories of humor~\citep{raskin1979semantic, forabosco1992cognitive} describe humor as arising from violated expectations followed by reinterpretation. Building on this view, computational studies have modeled incongruity through information-theoretic measures like \emph{surprisal}. Humorous headlines tend to have higher perplexities \citep{ijcai2021p0537}, and swapping low-probability (high surprisal) tokens with more predictable ones can remove humor \citep{horvitz-etal-2024-getting}. ~\citet{west2019reverse} further found that humor tends to reside in the later part of satirical headlines, aligning with the idea that humor emerges from late-stage expectation violation. In this section, we analyze the distribution of surprisal across sentence positions to test whether \textbf{Bulwer-Lytton sentences exhibit different patterns of incongruity} from standard humor datasets.

\paragraph{High surprisal tokens} We extract the log-probabilities for each token in our datasets using the \texttt{minicons} package~\citep{misra2022minicons} and the \texttt{gemma-3-1b-pt} LM~\citep{team2025gemma}. We define high surprisal tokens as those outside the top 1000 tokens ranked by probability as predicted by the LM, and plot their relative rank (predicted rank normalized by the vocabulary size) over relative sentence position for that token in its sentence in Figure~\ref{fig:surprisal}. We sample 1000 sentences from novel-openings and use the entirety of all other datasets. 

High-surprisal tokens from BL sentences differ in two ways from sentences in combo-humor. Firstly, high-surprisal tokens from combo-humor occur more often in the beginning half of the sentence, with a small peak at the end; BL sentences have a much flatter distribution of surprising tokens. Secondly, BL sentences have \textbf{more surprising tokens per sentence} (5.1$\pm$3.1) than combo-humor (2.0$\pm$1.4). Both GPT-5 and DeepSeek inflate the number of high-surprisal tokens, but curiously DeepSeek (6.2$\pm$2.6) does this less than GPT-5 (9.2$\pm$2.8). Figure~\ref{fig:surprisal} shows that DeepSeek's distribution aligns much more closely with that of human-written BL sentences than with GPT-5, which is relatively flat. The difference in behavior of these two LLMs adds to contemporaneous work demonstrating how different LLMs exhibit significant differences in behavior on the same task/prompt due to their distinct training pipelines~\citep{shao2025spuriousrewardsrethinkingtraining}. Figure~\ref{fig:app-surprisal} in Appendix~\ref{app:ans} shows the distribution of high-surprisal tokens for GPT-4.1 and \texttt{gpt-oss-120b}.

\begin{figure}
    \centering
    \includegraphics[width=\linewidth]{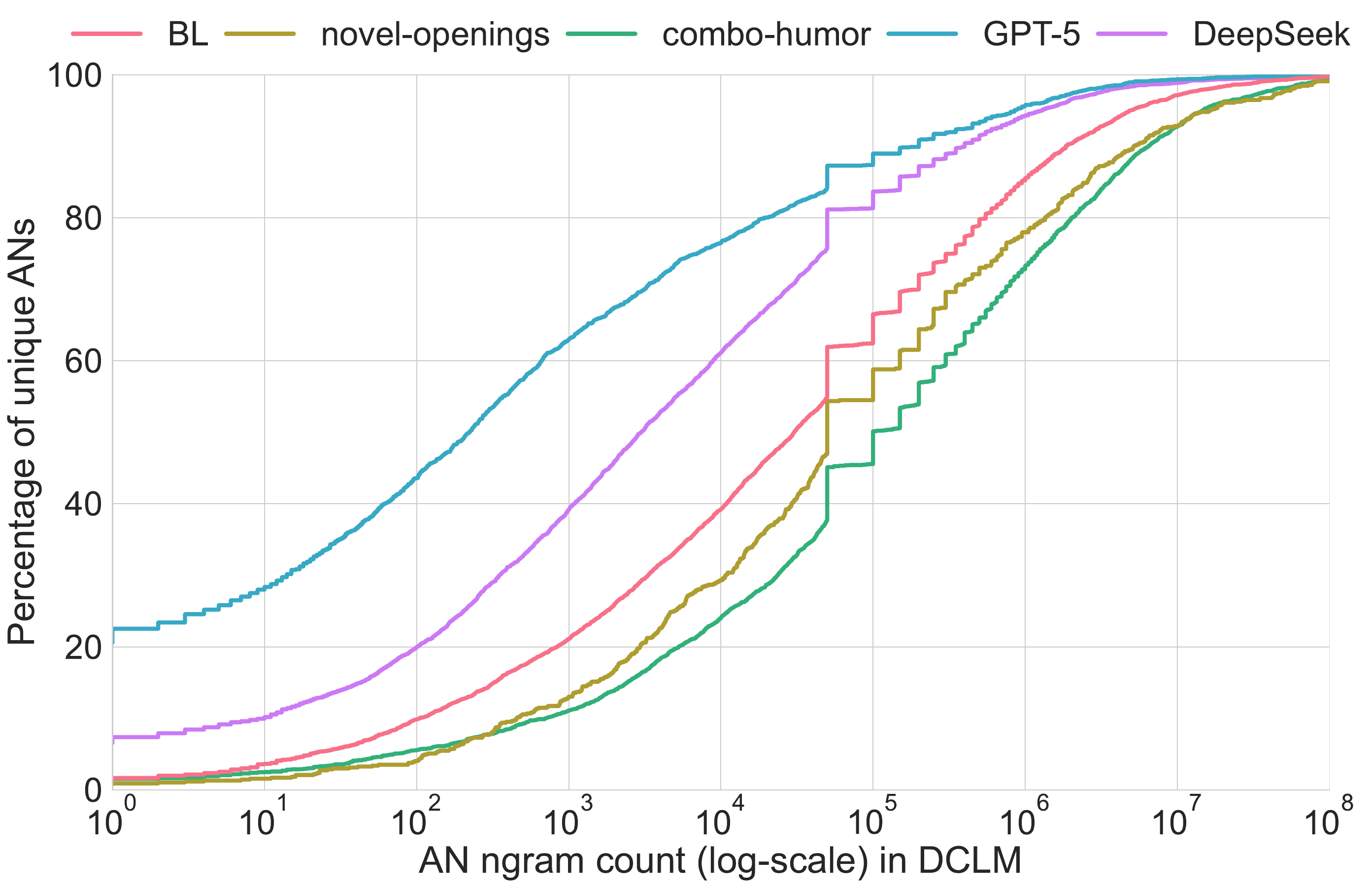}
    \caption{Cumulative distribution of percentage of ANs within a dataset against their count in the DCLM corpus.}
    \label{fig:ans}
    % \vspace{-\baselineskip}
\end{figure}

\begin{table*}[ht!]
    \centering
    \small
    \begin{tabular}{lp{0.8\linewidth}}
        \toprule
         BL & ocherous brilliance, albino apricots, helpless patrolman, indigent zucchini, antediluvian passion, crepuscular finish, tasty blowfly, dormie night\\\midrule
         combo-humor & underdressed layman, unfaithful kermit, nihilistic pencil, broke stroker, teenage fireflies, empathetic bovine, watchful potato\\\midrule
         novel-openings & ghastly armoire, legged lupine, mighty arrowhead, brassy tocsin, mean loudmouths\\\midrule
         GPT-5 & brocadeed accordion, tragic omelet, moonless noon, anti-gravity mustache, overripe chandelier, velveteen hush, moral drizzle, unblinking monocle\\\midrule
         GPT-4.1 & cosmic cheeseboard, narcoleptic willows, dimest shadows, filmy crinolines, deafening falcon, gelatinous moons \\\midrule
         DeepSeek & dubious grog, ghostly quadrille, enchanted paperweight, assertive sculpting, feisty seamstress, untrustworthy camels\\\midrule
         gpt-oss-120b & emberkissed moths, amber–wreathed memories, unholy expedition, syncopated agony, soggy rumors, olated secrets\\
         \bottomrule
    \end{tabular}
    \caption{Adjective-noun bigrams from our datasets that occur fewer than 10 times in the DCLM corpus.}
    \label{tab:ans}
\end{table*}

\subsection{Novel Adjective-Noun expressions} 
\label{subsec:ans}

Jarring and novel adjective–noun combinations disrupt semantic expectations and draw attention to the prose itself~\citep{westbury2019wriggly}, heightening the bad humor exemplified by BL sentences. Inspired by ~\citeposs{vecchi2017spicy} work showing that the `deviance' of novel adjective-noun pairs can be predicted by their distributional occurrence a corpus, we analyze the distribution of Adjective-Noun (AN) bigrams across all our datasets and their occurrence in a large pre-training corpus of natural language.

\paragraph{AN counts} As in \citet{vecchi2017spicy}, `deviant' or novel ANs are identified as ones that have a low count in a natural language corpus. We expect BL sentences to contain more novel AN bigrams than our baseline datasets. We extract all ANs from all our datasets (we sampled 1000 sentences for novel-openings) using Stanza~\citep{qi2020stanza} and query their count in the DCLM pre-training corpus~\citep{NEURIPS2024_19e4ea30} using Infini-gram~\citep{liu2024infinigram}.  BL sentences have more ANs per sentence on average (2.7) than novel-openings (0.77) or combo-humor (0.55), highlighting how multiple ANs are a unique feature for BL sentences. GPT-5 BL sentences had 2.56 ANs per sentence and DeepSeek sentences had 5.3, once again showing a curious difference in behavior from different LLMs to the same prompt that we also observed in Section~\ref{subsec:surprisal}. 

Figure~\ref{fig:ans} plots the cumulative percentage of all unique ANs in each dataset as a function of their (log-scaled) frequency in the DCLM corpus, allowing us to see, for any frequency threshold, what share of ANs occur at or below that count. As expected, BL sentences have many more novel ANs per sentence than either of our baseline datasets --- 10\% of BL ANs occur less than 100 times in DCLM compared to 4\% and 6\% for novel-openings and combo-humor. 

Furthermore, Figure~\ref{fig:ans} (and Figure~\ref{fig:app-ans} in Appendix~\ref{app:ans}) shows that synthetic BL sentences have even more novel ANs than the human-written BL sentences. Together with our analysis in previous sections, this paints a consistent picture of the synthetic BL sentences \textbf{imitating the form of human-written BL sentences while consistently exaggerating its notable features}. Table~\ref{tab:ans} showcases examples of novel ANs that occur fewer than 10 times in the DCLM corpus from our datasets.

\section{Conclusion}
\label{sec:conclusion}
We introduce a new dataset of intentionally bad humor with sentences from the Bulwer-Lytton Fiction Contest. Humor detection models fail to identify these sentences as humorous, and literary theme analysis further reveals how devices like metaphor, repetition, and metafiction differentiate the humor in this dataset. Analysis of synthetic BL sentences reveals that they imitate the form of BL sentences, but exaggerate several of their key features.

\section*{Limitations}

\paragraph{Data and focus} Our work is a preliminary investigation into `bad' humor as exemplified by entries in the Bulwer Lytton Fiction Contest, and we wish to motivate how our novel corpus differs from existing humor datasets. Intentionally bad humor can occur in many forms --- sentences in the BLFC are simply one form of that. Whether or not a sentence constitutes bad humor or a good entry in the BLFC is an inherently subjective question. We leave the quantification and further analysis of the boundaries of when humor is `bad' or `good' to future work, and rely on the subjective choices by the organizers of the BLFC.

\paragraph{Humor Detection} Our experiments use a limited number of humor detection models; it is possible that other models would perform better on out-of-domain data. However, we believe that the complexity of BL sentences and their tendency to encapsulate multiple forms of humor means that other models would also struggle to quantify their humor.

\paragraph{Literary Device Analysis} We focus on only eight key literary devices identified using TopicGPT with GPT-4.1. In preliminary experiments, we found that other LLMs identified additional literary devices such as personification, but also identified false positives (e.g., the inclusion of ``Craigslist'' in the sentence). Following a pilot of the intruder task discussed in Appendix~\ref{app:topics}, we decided that the literary devices generated and assigned using GPT-4.1 gave us the best overview of the dataset while excluding noisy features. Despite this, a small number of sentences were labeled by GPT-4.1 in a way that could not be correctly parsed during the assignment phase. In some cases, a feature was assigned with the qualifier (``not assigned'') in the description of why it was assigned, and in other cases features were assigned that did not actually exist in our set of eight literary devices. This introduces a small amount of noise, but our feature validation process shows that the labels are still meaningful.

\paragraph{Synthetic BL sentences} Our work focuses on understanding how LLMs generate and mimic the style of BLFC entries with minimal prompting --- hence the one-shot example with simple instructions from the rules on the website. We focus on how these sentences differ stylistically from the human-written BL sentences in this paper, and leave evaluation of whether the generated sentences are good entries for the BLFC to future work.

\paragraph{Surprisal} While we analyze the incongruity in BL sentences through token probabilities, this does not imply \emph{directly} that the humor resides in regions of high surprisal. This would require human annotation and further analysis (as in \citet{west2019reverse}) which goes beyond the scope of this short paper. Our goal with the analysis in this paper is to motivate that the distribution (and number) of low-probability tokens is clearly different between BL sentences and standard sentences of sentential humor, as well as between human-written and synthetic BL sentences.

\section*{Acknowledgments}
This research was partially supported by start-up funds and computational resources provided by Ithaca College and Middlebury College. We would also like to thank Kasia Bartoszynska and Hugh Egan in the English department at Ithaca College for helpful feedback and discussions in the early stages of this work.  Finally, we would like to thank the six Middlebury College undergraduate students who helped with data annotation. AI assistants (Gemini, Claude Code) were used for writing some of the code used in experiments, and for editing/re-writing sentences in the final paper.

% Bibliography
\bibliography{references}

\appendix
\label{sec:appendix}
\section{Example BL sentences}
\label{app:bl-exs}
The opening sentence from the novel \emph{Paul Clifford} by Edward Bulwer-Lytton that inspired the contest:

\ex. It was a dark and stormy night; the rain fell in torrents---except at occasional intervals, when it was checked by a violent gust of wind which swept up the streets (for it is in London that our scene lies), rattling along the housetops, and fiercely agitating the scanty flame of the lamps that struggled against the darkness.

Below, we list additional examples from the Bulwer-Lytton fiction contest:

% Here are some entries from the Bulwer-Lytton fiction contest:

\ex. \a. Little Timmy suffered from Claustraphobia: the fear of being trapped in a closet with Santa Claus.
\b. She had a body that reached out and slapped my face like a five-pound ham-hock tossed from a speeding truck. 
\b. Space Fleet Commander Brad Brad sat in silence, surrounded by a slowly dissipating cloud of smoke, maintaining the same forlorn frown that had been fixed upon his face since he’d accidentally destroyed the phenomenon known as time, thirteen inches ago.
\b. Having committed to memory the route maps leading from Alaska to L.A. and determined to avenge the more than 300 years of ridicule his people had suffered for the popularly assumed whimsy of Eskimo naming conventions, Robson Pollowow ("He who moves mountains") glanced one last time into his hastily rigged rear view mirror before releasing the hand brake on the Paw Whal Dee Glacier.

\section{One-shot prompt}
\label{app:prompt}

\begin{prompt}[title={Prompt for generating BL sentences},label=oneshot-prompt]
The Bulwer-Lytton Fiction Contest challenges participants to write an atrocious opening sentence to the worst novel never written. The whimsical literary competition honors the novelist Sir Edward George Bulwer-Lytton and his marvelously awful opening to his 1830 novel Paul Clifford:\\

``It was a dark and stormy night; the rain fell in torrents—except at occasional intervals, when it was checked by a violent gust of wind which swept up the streets (for it is in London that our scene lies), rattling along the housetops, and fiercely agitating the scanty flame of the lamps that struggled against the darkness.''\\

The rules for the Bulwer Lytton Fiction Contest are childishly simple:
\begin{itemize}
    \item Each entry must consist of a single sentence.
    \item Sentences may be of any length but we strongly recommend that entries not go beyond 50 or 60 words.
    \item Entries must be "original" (as it were) and previously unpublished.
\end{itemize}

Based upon these instructions and the original example, your goal is to write the most atrocious opening sentence to the worst novel ever written in the following genre: \{genre\}.  Your final output should contain only the sentence, with no other text or explanations.
\end{prompt}

We sample from the following genres which were the most popular in the human-written BL dataset: Adventure, Science Fiction (100 each), Purple Prose, Romance, Crime \& Detective, Vile Puns (150 each), Western, Historical Fiction, Children's Literature, and Fantasy \& Horror (50 each). The numbers for each genre were chosen because they closely followed the relative proportion of sentences from each genre in the human-written BL corpus.

We generate samples from GPT-4.1 and GPT-5 using the OpenAI API (\url{https://platform.openai.com}), and DeepSeek using the Together API (\url{https://api.together.ai}). Samples from \texttt{gpt-oss-120b} were generated locally on a workstation with 2 Nvidia RTX 6000 Ada GPUs. All responses were generated with temperature set to 1 and with thinking set to off (or low).

\section{Literary Devices}
\label{app:topics}

\begin{figure}[ht!]
    \centering
    \includegraphics[width=\linewidth]{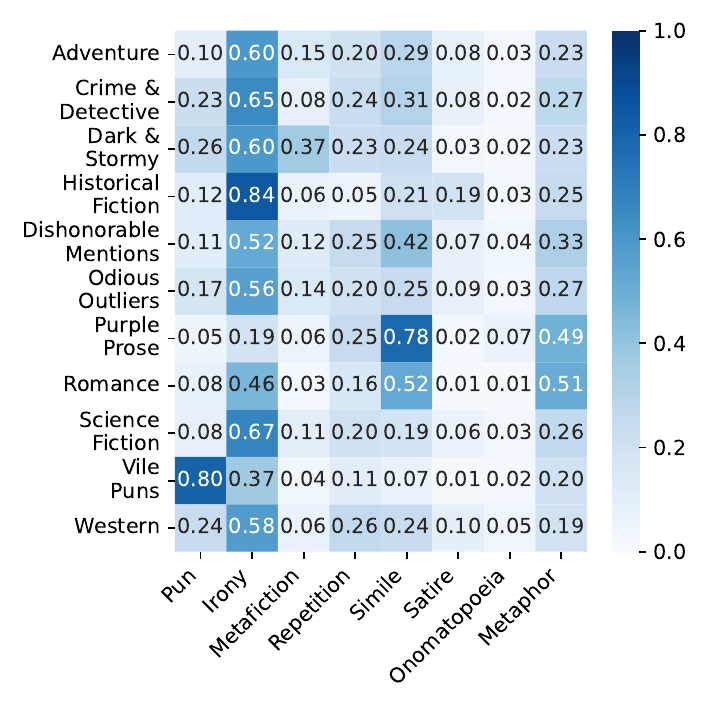}
    \caption{Literary device presence in BL sentences from different categories.}
    \label{fig:topic-by-category}
\end{figure}

\begin{figure}[ht!]
    \centering
    \includegraphics[width=\linewidth]{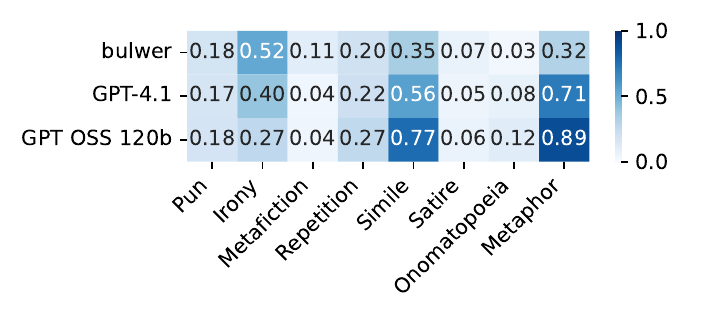}
    \caption{Literary device presence in BL sentences compared to GPT-4.1 and \texttt{gpt-oss-120b}.}
    \label{fig:topic-v-generated}
\end{figure}

Figure~\ref{fig:topic-by-category} shows the literary device breakdown by category, focusing on those categories linked to at least 50 entries--for instance, 80\% of entries categorized as ``Vile Puns'' were assigned the Puns feature and 78\% of entries categorized as ``Purple Prose'' were assigned the Simile feature. Figure~\ref{fig:topic-v-generated} shows the literary device breakdown for two additional models used to generate synthetic BL sentences, GPT-4.1 and gpt-oss 120b. We find that like with the other synthetic sentences, simile, metaphor, and onomatopoeia are exaggerated.

\paragraph{Feature Validation}
We conduct an intruder task similar to the task introduced by \citet{chang2009reading} to evaluate the quality of the feature evaluations. Human annotators are shown a feature, its description, and five BL sentences. One sentence was not assigned the feature and the other four were; the annotator must determine which is the ``intruder.''

\begin{prompt}[title={Instructions Given to Annotators},label=annotation-instructions]
\textbf{Background}\\
We are working on a project to characterize the type of humor in winners of the Bulwer-Lytton fiction contest and how it differs from other humor datasets. We have extracted a set of literary features for each sentence. The task is intended to determine the quality of the extracted features.\\
\textbf{Your Task}\\
You are going to see a page with a question (each question is for a specific literary feature) and five examples labeled A-E. In each case, there are going to be four positive examples that were labeled with that feature by a model and one negative example that was not labeled with that feature. For each question, you should select the answer corresponding to the example that you think is least representative of the feature (so you're identifying the negative example). This is a situation where you just have to do your best. It's possible there will be multiple examples that don't fit well or that all of the examples will fit (which indicates the model isn't doing a good job) so select whichever makes the most sense to you! Read over the sentences carefully, as they are fairly complex.
\end{prompt}

We created six instances for each of the eight features, for a total of 48 unique instances. Six annotators were randomly assigned 16 instances each, and each instance was annotated by two annotators.\footnote{All annotators were undergraduate students and were given a \$15 gift card for approximately one hour of work.} One would expect 20\% accuracy if the feature assignments did not match the text and annotators always guessed randomly. Among the annotators, accuracy ranged from $\frac{7}{16}$--$\frac{12}{16}$, with an average of 57\% of instances labeled correctly.

Using Krippendorff's alpha, we found that annotators had fairly low agreement (0.51). This is unsurprising given that the sentences are fairly complex and difficult to parse. However, we believe that the accuracy indicates the usefulness of the features despite the difficulty of the annotation task.

\section{Deviant ANs \& Surprisal}
\label{app:ans}

\begin{figure}[H]
    \centering
    \includegraphics[width=\linewidth]{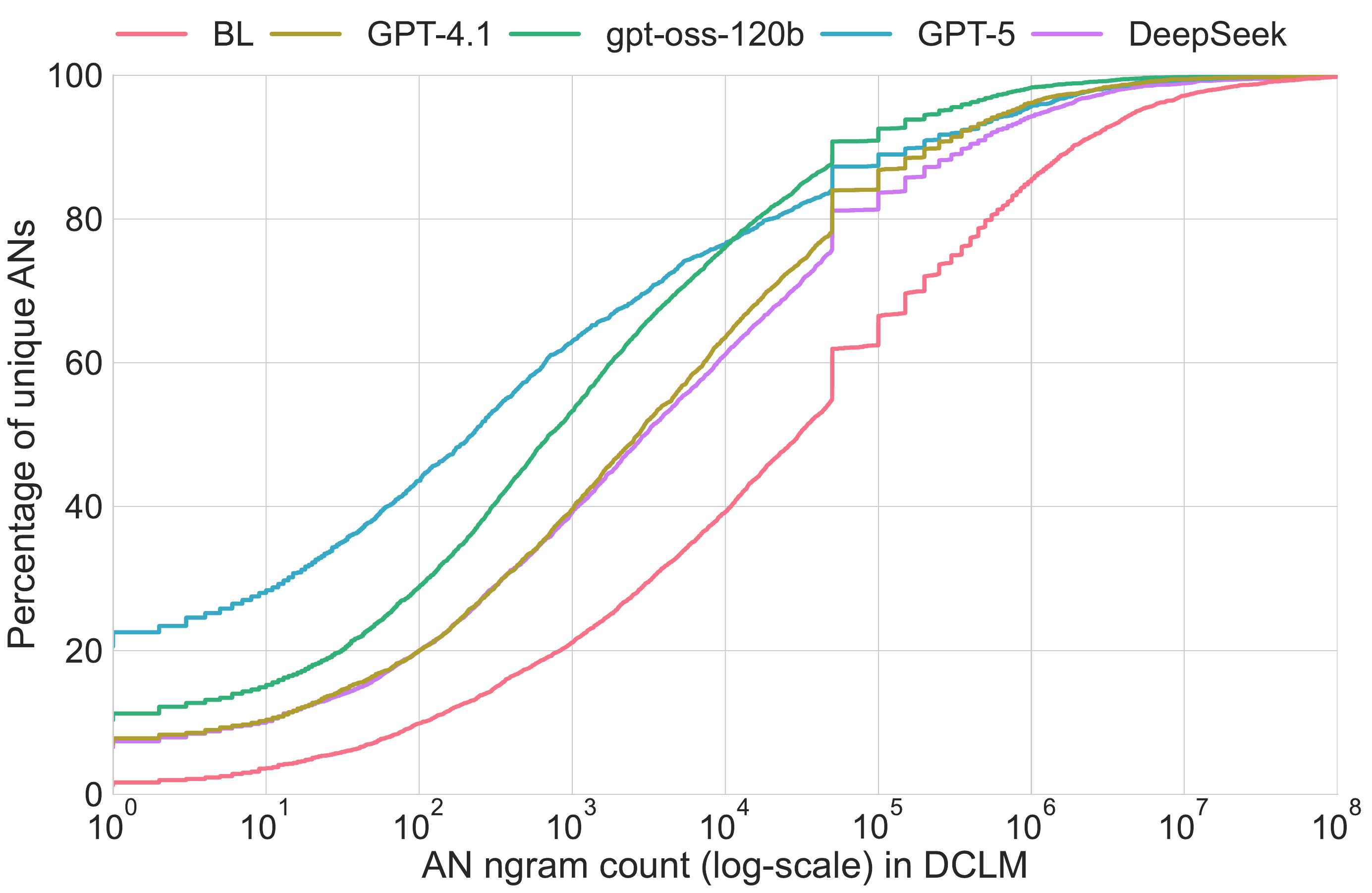}
    \caption{Cumulative distribution of percentage of ANs within a dataset against their count in the DCLM corpus.}
    \label{fig:app-ans}
\end{figure}

When querying the count for ANs against the Infinigram API, we include a disjunction over lower-cased and capitalized versions of the AN expression. Figure~\ref{fig:app-ans} plots the cumulative percentage of all unique ANs in the synthetic and human-written BL sentences as a function of their (log-scaled) frequency in the DCLM corpus: GPT-4 and DeepSeek exhibit similar distributions, while \texttt{gpt-oss-120b} and GPT5 differ significantly as they generate many more unattested ANs. Table~\ref{tab:ans} shows some rare ANs from each of our datasets.

\begin{figure*}[t]
    \centering
    \includegraphics[width=\textwidth]{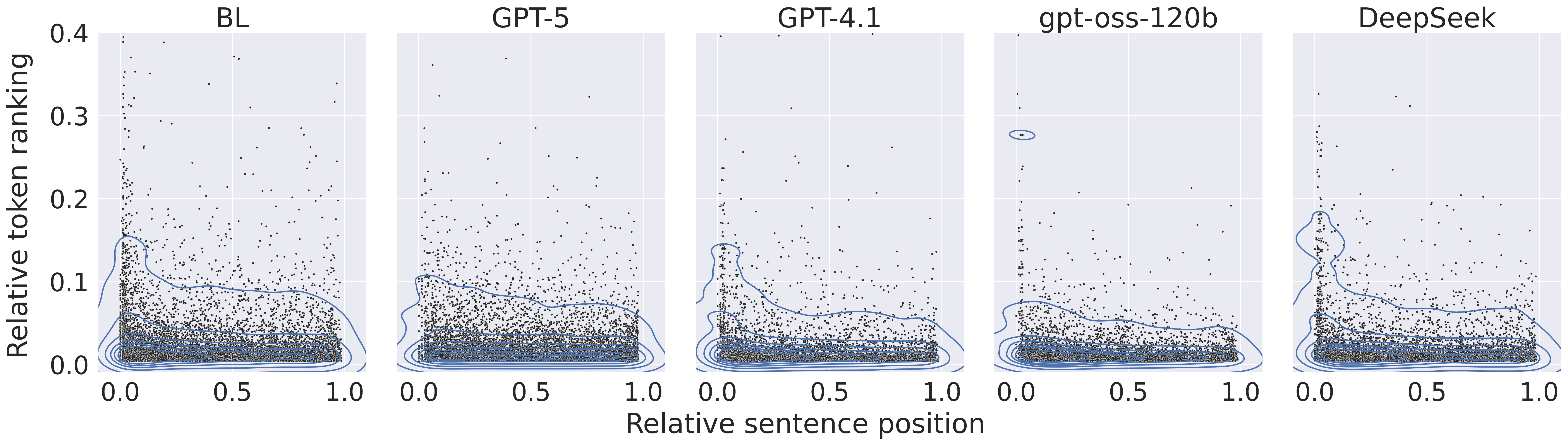}
    \caption{Relative rank of high-surprisal tokens plotted against their relative sentence position for \texttt{gpt-oss-120b} and \texttt{GPT-4.1}.}
    \label{fig:app-surprisal}
\end{figure*}

\end{document}